\documentclass[]{spie}  

\usepackage{multirow}

\usepackage{float}
\usepackage{amsmath,amsfonts,amssymb, bm}
\usepackage{graphicx}
\usepackage{adjustbox}
\usepackage[colorlinks=true, allcolors=blue]{hyperref}
\usepackage{xcolor}
\usepackage{booktabs}
\usepackage{amsmath,xcolor}
\usepackage{mathtools}
\usepackage{hhline}

\title{VTR: An Optimized \underline{V}ision Transformer for SAR A\underline{TR} Acceleration on FPGA }

\author[a]{Sachini Wickramasinghe*}
\author[a]{Dhruv Parikh*}
\author[a]{Bingyi Zhang}
\author[b]{Rajgopal Kannan}
\author[a]{\\Viktor Prasanna}
\author[b]{Carl Busart}
\affil[a]{University of Southern California, Los Angeles, CA}
\affil[b]{DEVCOM Army Research Office, Playa Vista, CA}

\authorinfo{Further author information: (Send correspondence to Sachini Wickramasinghe)\\Sachini Wickramasinghe: shwickra@usc.edu
\\Dhruv Parikh: dhruvash@usc.edu
\\Bingyi Zhang: bingyizh@usc.edu
\\Rajgopal Kannan: rajgopal.kannan.civ@army.mil
\\Viktor Prasanna: prasanna@usc.edu
\\Carl Busart: carl.e.busart.civ@army.mil}

\begin{document} 
\maketitle







\footnotetext{*: Equal contribution}
\begin{abstract}
Synthetic Aperture Radar (SAR) Automatic Target Recognition (ATR) is a key technique used in military applications like remote-sensing image recognition. Vision Transformers (ViTs) are the state-of-the-art in various computer vision applications, outperforming Convolutional Neural Networks (CNNs). However, using ViTs for SAR ATR applications is challenging due to (1) standard ViTs require extensive training data to generalize well due to their low locality. The standard SAR datasets have a limited number of labeled training data, reducing the learning capability of ViTs (2) ViTs have a high parameter count and are computation intensive which makes their deployment on resource-constrained SAR platforms difficult. In this work, we develop a lightweight ViT model that can be trained directly on small datasets without pre-training. To this end, we incorporate the Shifted Patch Tokenization (SPT) and Locality Self-Attention (LSA) modules into the ViT model. We directly train this model on SAR datasets to evaluate its effectiveness for SAR ATR applications. The proposed model, VTR (\underline{V}iT for SAR A\underline{TR}), is evaluated on three widely used SAR datasets: MSTAR, SynthWakeSAR, and GBSAR. Experimental results show that the proposed VTR model achieves a classification accuracy of 95.96\%, 93.47\%, and 99.46\% on MSTAR, SynthWakeSAR, and GBSAR datasets, respectively. VTR achieves accuracy comparable to the state-of-the-art models on MSTAR and GBSAR datasets with $1.1\times$ and $36\times$ smaller model sizes, respectively. On SynthWakeSAR dataset, VTR achieves a higher accuracy with a model size that is $17\times$ smaller. Further, a novel FPGA accelerator is proposed for VTR, to enable real-time SAR ATR applications. Compared with the implementation of VTR on state-of-the-art CPU and GPU platforms, our FPGA implementation achieves latency reduction by a factor of $70\times$ and $30\times$, respectively. For inference on small batch sizes, our FPGA implementation achieves a $2\times$ higher throughput compared with GPU.


\end{abstract}
 

\section{INTRODUCTION}
\label{sec:intro}  
Automatic Target Recognition (ATR) for Synthetic Aperture Radar (SAR) images is a broadly researched topic due to its diverse applications spanning from remote sensing to military surveillance \cite{TSOKAS2022117342}. Unlike optical sensors, SAR can capture high-resolution images irrespective of weather conditions or time (day/night). Recent advances in SAR imaging systems have led to images with resolutions as high as few decimeters \cite{6355595_veryhighressar}. These advances, coupled with their any-circumstance imaging capability, have led to SAR based sensor systems outperforming their optical sensor system counterparts, while being more reliable. ATR for SAR comprises of three distinct tasks: detection, discrimination and classification \cite{rs15051454}. Detection involves identifying regions-of-interest within a SAR image to localize targets. Discrimination involves the capability of an ATR algorithm to discard false alarms generated by the detection algorithms due to natural/artificial clutter. Classification involves accurately and precisely classifying detected targets within a SAR image \cite{rs15051454}. Since the images generated by radar sensors differ significantly from the images generated by optical sensors \cite{6504845_sartut}, SAR ATR becomes a challenging problem to solve. 

Recent advances in deep learning have revolutionized the field of SAR ATR \cite{rs15051454}. Several works employing deep learning to SAR ATR utilize traditional Convolutional Neural Networks (CNNs) \cite{7393462_cnn_1, 7460942_cnn_2} to classify SAR images, outperforming prior works. Graph Neural Network (GNN) based approaches \cite{zhang2023graph, 9632598_dgnn_sar} have led to state-of-the-art performance for SAR ATR applications. GNNs drastically reduce the overall model parameters and inference latency, making them suitable for real-time applications. Vision Transformer (ViT), introduced by Dosovitskiy et al. \cite{dosovitskiy2021image}, has outperformed traditional CNN based architectures across several tasks in the computer vision domain \cite{Khan_2022}. Several recent works \cite{chen2022geospatial, liu2022high, wang2022global} have employed ViTs across various SAR ATR applications. Chen et al. \cite{chen2022geospatial} utilize a multi-scale geospatial contextual attention network (MGCAN) on several SAR image chips, inspired from the Multi-Head Self-Attention mechanism in ViTs. MGCAN model was used in Chen et al. \cite{chen2022geospatial} to perform object detection for aircrafts. Liu et al. \cite{liu2022high} utilize ViTs and CNNs to extract global and local features, respectively. The combined model led to an improved classification accuracy when employed for image classification on high resolution (HR) SAR images. Wang et al. \cite{wang2022global} also combine ViTs and CNNs into a module termed ConvT. ConvT was utilized for few shot image classification on small-sized SAR datasets.

Despite ViTs being the state-of-the-art model for vision applications, significant challenges prevent its effective deployment for SAR ATR applications. (i) SAR ATR datasets are typically quite small. SAR image collection is an expensive endeavour. Thus, most available SAR datasets have limited number of training instances. ViTs generally require a large amount of training data in order to generalize due to its limited locality inductive bias \cite{dosovitskiy2021image}. Thus, training a raw ViT on the small-sized SAR datasets without pre-training on larger datasets becomes challenging. (ii) ViTs are computationally expensive with a large memory footprint and associated model size. Since SAR ATR applications are driven by real-time constraints, it becomes imperative to optimize trained models for efficient real-time inference. The computational cost of ViTs is proportional to the square of the total input tokens \cite{dosovitskiy2021image}. For images with higher resolutions (such as SAR), this leads to an intractably high computation cost.

Prior works utilize ViTs pre-trained on large datasets such as ImageNet \cite{5206848_imgnt}. The pre-trained ViTs are then finetuned on small SAR ATR datasets \cite{9658539_image_cls, liu2022high, 9947220_PVT_SAR}. Further, while several works accelerate SAR ATR applications \cite{10035150_bingyi_sar_atr, 10363615_bingyi_sar_atr_hbm, fein2024single} on FPGA, these works do not focus on optimizing the ViT architecture for SAR ATR.

To address the above challenges, we propose VTR, a novel \underline{V}iT based model for SAR A\underline{TR} application. VTR can be trained directly on the small-sized SAR datasets, without pre-training on larger datasets. Furthermore, we propose a novel FPGA accelerator for low-latency and high-throughput SAR ATR. Our contributions are as follows,

\begin{itemize}
    \item We propose a novel ViT model (VTR) equipped with the Shifted Patch Tokenization (SPT) and Locality Self-Attention (LSA) modules \cite{lee2021vision} for SAR ATR. VTR improves the locality inductive bias of ViTs, allowing it to directly be trained on small SAR datasets, removing the overhead associated with pre-training.
    \item We propose a novel accelerator for the proposed VTR model on FPGA. We propose a Highly Parallel Processing Unit (HPPU) to fully exploit the compute parallelism within each layer (encoder) of a VTR.
    \item We comprehensively evaluate the performance of several configurations of VTR across the MSTAR \cite{MSTAR}, SynthWakeSAR \cite{rizaev2022synthwakesar_synthwakesar} and GBSAR \cite{10133345_gbsar} datasets. The performance of a model is characterized via classification accuracy, model size, computation complexity, and inference latency and throughput. 
    \item VTR achieves a higher classification accuracy for SynthWakeSAR dataset with a $17\times$ smaller model size. On MSTAR and GBSAR datasets, VTR achieves an accuracy comparable to the current state-of-the-art, with a $1.1\times$ and $36\times$ smaller model size, respectively. 
    \item The proposed VTR FPGA accelerator reaches a latency reduction of $30\times$ and $70\times$ compared to state-of-the-art GPU and CPU platforms, respectively. For inference on small batch sizes, it reaches a throughput improvement of $2\times$ when compared with GPU.
\end{itemize}




\section{BACKGROUND AND RELATED WORK}
\subsection{Deep Learning models for SAR ATR}
Deep neural networks have gained high interest, showcasing impressive results across various problem domains. The state-of-the-art works for SAR ATR applications involve either convolutional neural networks (CNNs) or Graph Neural Networks (GNNs)\cite{fein2023benchmarking}. Morgan \cite{morgan2015deep} is the first work to propose a deep CNN for SAR ATR. Recently, Zhang et al.\cite{zhang2023graph} proposed a novel architecture based on GNNs that exceeded classification accuracy of 99\% on the MSTAR dataset. In contrast to CNNs, the proposed GNN exploits the data sparsity in SAR images to reduce computation complexity. Although ViTs have emerged as state-of-the-art models for computer vision tasks, they perform poorly on small datasets due to severe overfitting. Hence, training a ViT model for SAR ATR applications without pre-training is challenging. Li et al.\cite{li2021automatic} proposed a pre-processing technique to improve the accuracy of the ViT model on SAR image classification. Recently, Shifted Patch Tokenization (SPT) and Locality Self-Attention (LSA) techniques have been proposed to improve the accuracy of ViTs on small datasets\cite{lee2021vision}. In this work, we propose VTR model for SAR ATR, which incorporates the SPT and LSA modules into a vanilla ViT model. This enables training VTR on small SAR datasets without pretraining.

\subsection{Vision Transformer}
ViT\cite{dosovitskiy2021image} has achieved significant advancements in computer vision tasks like image classification and object detection. Its core components include Multi-Headed Self-Attention (MSA) and Multi-Layer Perceptron (MLP) blocks. First, the input image undergoes partitioning into non-overlapping patches. Then, the flattened patches are embedded along with the class token and positional information. Subsequently, this processed data is fed into the transformer encoder. \\\\
\textbf{Multi-Headed Self-Attention:} In the self attention block, the input embeddings are linearly projected to query, key, and value vectors. The query and key vectors are then utilized to obtain a scaled dot product. A single self-attention operation is denoted as one "head". The self-attention function is defined as,
\begin{equation}
\text{Attention}(\bm{Q},\bm{K},\bm{V}) = \text{softmax}(\frac{\bm{Q}\bm{K}^T}{\sqrt{d_k}})\bm{V}
\label{eq: _attn__}
\end{equation}
where $\bm{Q}$, $\bm{K}$ and $\bm{V}$ represent the query, key, and value matrix, respectively. $d_k$ is the dimension of embeddings in $\bm{K}$. Several such heads are concatenated and projected via a linear layer to compute the final MSA output. \\\\
\textbf{Multi-Layer Perceptron:} The output of the MSA layer is fed into the MLP block. It consists of two linear layers with an activation function. The MLP layer is formulated as,\\
\begin{equation}
\text{MLP}(\bm{x}) = \text{GeLU}(\bm{x} \bm{W}_1+\bm{b}_1)\bm{W}_2+\bm{b}_2\\
\end{equation}
where $\bm{W}_1$ and $\bm{b}_1$ represent the hidden layer weights and bias, respectively. $\bm{W}_2$ and $\bm{b}_2$ are the output layer weights and bias respectively. GeLU is the activation function.\\\\
 In the standard ViT model, the receptive field of the input embeddings (input tokens) remains fixed and cannot be adjusted. That is, the tokenization of the standard ViT is similar to the operation of a non-overlapping convolutional layer\cite{lee2021vision}. As a result, these tokens have a small receptive field leading to lower local inductive bias. To this end, SPT leverages spatial information by shifting the input image in the four or eight cardinal directions. Additionally, since both the query and key are linearly projected from the same input tokens, the self-token relations tend to exhibit larger magnitudes compared to inter-token relations. Consequently, the softmax function assigns relatively higher scores to self-token relations and smaller scores to inter-token relations. Hence, the attention of standard ViT tends to be similar to each other regardless of inter-token relations. Moreover, the scaling factor $\sqrt{d_k}$ can cause smoothing of the attention score distribution\cite{he2018determining}. These issues can be effectively mitigated by incorporating LSA into the attention layer. LSA helps in excluding self-token relations and applies a learnable temperature scale to the softmax function.

\section{Overview}
\label{subsec:sys_overview}

\begin{figure}[ht]
\centering
\includegraphics[scale=0.4]{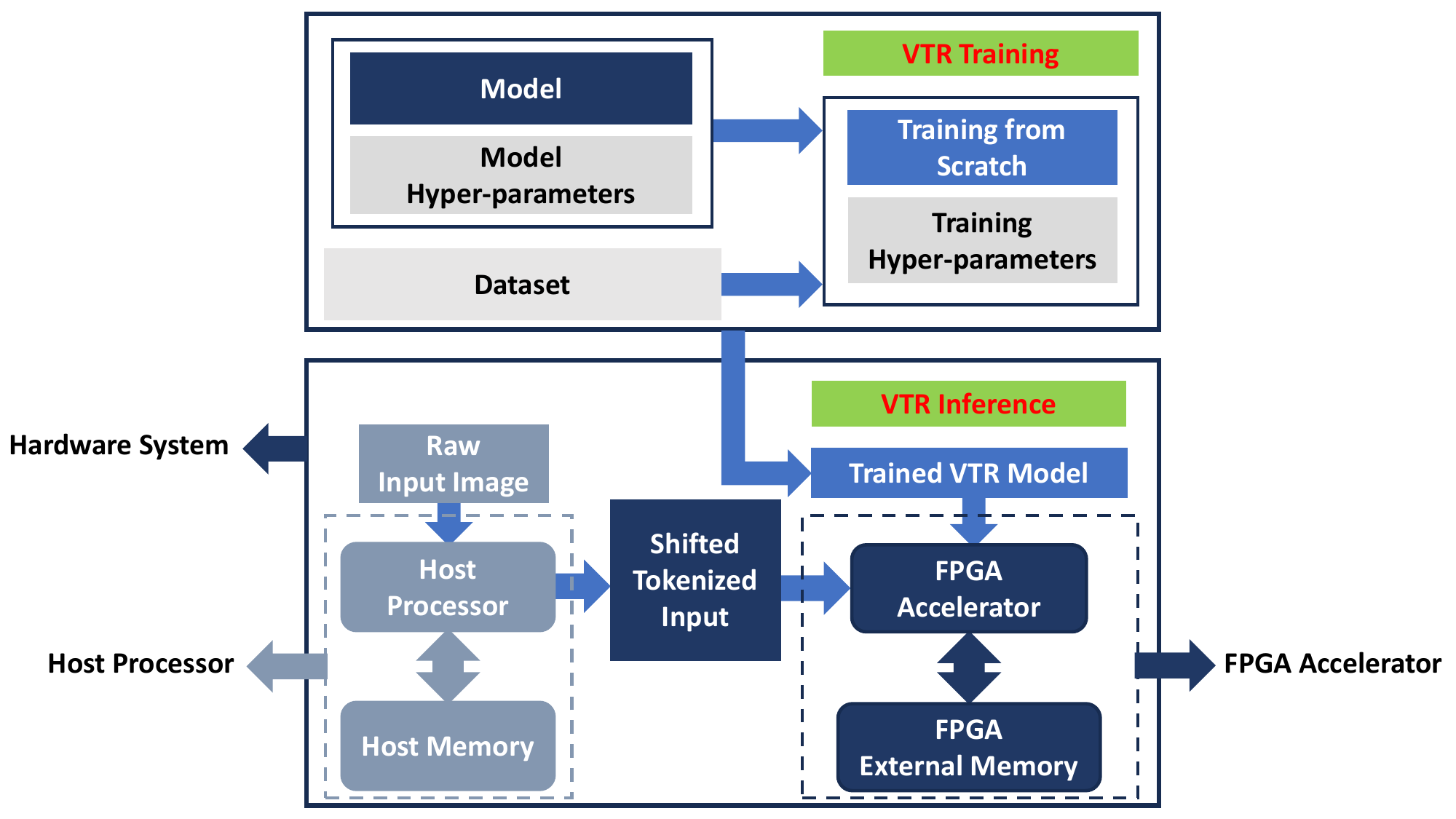}
\caption{Overview}
\label{fig:sys_ovw}
\end{figure}

The overview of the proposed approach is detailed in Figure \ref{fig:sys_ovw} and comprises of two stages: (i) VTR Training and (ii) VTR Inference. 
\vspace{-5mm}
\paragraph{VTR Training} The proposed VTR model (Section \ref{sec:Model_Des}) is trained on small SAR datasets, without pre-training. As such training is inexpensive, we train several VTR model variants by varying the model hyper-parameters, for each SAR dataset, to evaluate the performance of VTR model across each setting.

\vspace{-5mm}
\paragraph{VTR Inference} Next, the trained VTR model is accelerated on an FPGA using a novel accelerator proposed in Section \ref{subsec: hw_des}. The input to the accelerator is the shifted and tokenized input image - post the application of the image shifting, concatenation and tokenization action of the SPT module. Specifically, for an image $\bm{x} \in R^{H \times W \times C}$, the host processor shifts the input image along the four cardinal directions. Each such shifted image, along with the original raw image, are concatenated along the channel axis to generate a shifted concatenated image: $S(\bm{x}) \in R^{H \times W \times (N_s + 1) C}$. Here, $S(.)$ represents the shifting module and $N_{s}$ represents the number of shifted images. Finally, the shifted concatenated image $S(\bm{x})$ is tokenized by partitioning $S(\bm{x})$ into several patches and flattening each patch into a embedding vector to generate the final shifted tokenized input for the FPGA accelerator, $S(\bm{x}) \xrightarrow[]{\text{tokenize}} \bm{X}$, where $\bm{X} \in R^{N \times D}$ represents the tokenized input with $N$ tokens (total patches) and $D$ embedding dimension (size of a flattened patch).

\section{MODEL DESIGN}
\label{sec:Model_Des}
\begin{figure}[ht]
\centering
\includegraphics[scale=0.46]{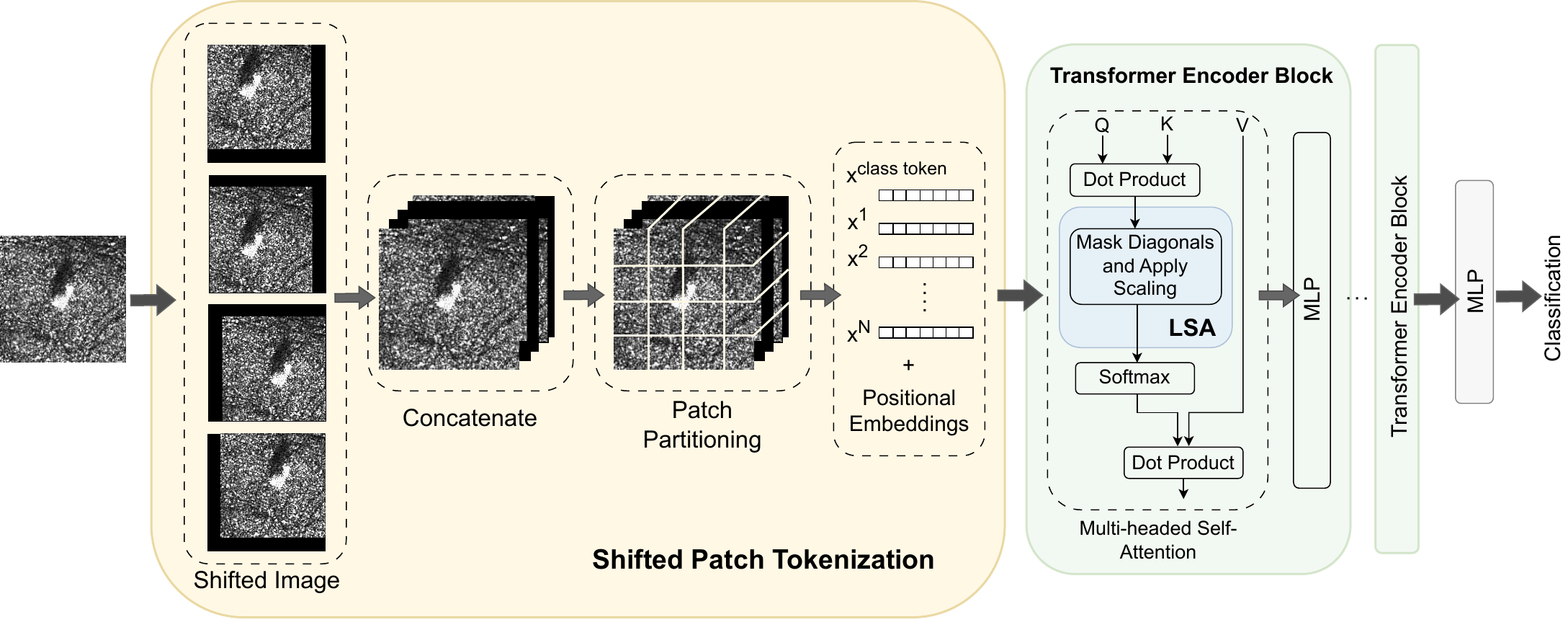}
\caption{Model Architecture}
\label{fig:model}
\end{figure}

The primary focus of our work is to implement a ViT that can effectively learn from small SAR datasets without any pre-training. In this Section, we present the architectural overview of the model. Then, we introduce the two modules: SPT and LSA\cite{lee2021vision}. The overall architecture is illustrated in Figure \ref{fig:model}.
\subsection{Overview of the Model}
Our model first transforms the input image using SPT. Subsequently, both the original input image and the transformed images (shown in Figure \ref{fig:model}) are concatenated and partitioned into non-overlapping patches. The flattened patches undergo layer normalization and linear projection to obtain patch embeddings which are concatenated with a learnable class embedding and then added to learnable positional embeddings to yield the final input tokens. This is then fed into the transformer encoder where the input is passed through several layers of multi-headed self-attention and multi-layer perceptron networks. In the multi-headed self-attention layers, our model incorporates the LSA module. The LSA module applies a learnable temperature scaling to the softmax function to sharpen the distribution of attention scores. Layer normalization is applied before every attention and MLP block while residual connections are applied after each block. Finally, the output of the transformer encoder is passed through an MLP head to generate the classification result.

\subsection{SPT}
In the SPT module, each input image, $\bm{x} \in \mathbb{R}^{H\times W\times C}$, is shifted by $2$ pixels in the four diagonal directions: left-up, right-up, left-down, and right-down. These shifted images are cropped to the same size as the original input image and concatenated with it (Figure \ref{fig:model}). The resulting set of images are then divided into $N$ non-overlapping patches and flattened into a sequence of vectors, $\bm{x}^i \in \mathbb{R}^{P^2.C.(N_s+1)}$. $H,W,$ and $C$ represent the height, width and the number of channels of the input image respectively. In our case, $C=1$. $\bm{x}^i$ is the $i^{th}$ flattened vector. $P$ indicates the size of the patch while $N_s$ represents the total number of shifted images. The patch embeddings (flattened vectors) are linearly transformed into a hidden dimension, $d_s$. A class token is concatenated to the linearly transformed patch embeddings and positional embeddings are added. The output of this SPT module is fed to the transformer encoder.

\subsection{LSA}
The main components of the LSA module include diagonal masking and learnable temperature scaling. In the multi-headed self-attention layer, two linear projections of the input are generated: the query, $\bm{Q} = \bm{X}\bm{W}_q$, and the key, $\bm{K} = \bm{X}\bm{W}_k$. As formulated in eq.\ref{eq: _attn__}, the softmax function is applied on a scaled dot product of these linear projections. The diagonal of the resultant matrix of the dot product, $\bm{Q}\bm{K}^T$, represents the self-token relations while the off-diagonal elements represent the inter-token relations. Since the linear projections are from the same input tokens, the self-token relations tend to be larger resulting in higher scores for self-token relations. To prevent this the LSA module forces $-\infty$ on the diagonal elements. This effectively prevents the attention from being focused on its own tokens. The learnable temperature scaling technique incorporated in the LSA module, allows the ViT to decide the softmax temperature by itself while training. 


\section{Accelerator Design}
\label{subsec: hw_des}

In this Section, we describe the designed hardware accelerator comprising of two core compute modules: (i) Highly Parallel Processing Unit (HPPU) and (ii) Element-wise Compute Unit (ECU). Prior to their description, we briefly discuss the data layout (format) for the accelerator.

\subsection{Data Layout}
\label{subsubsec: layout}

\begin{figure}[ht]
\centering
\includegraphics[scale=0.4]{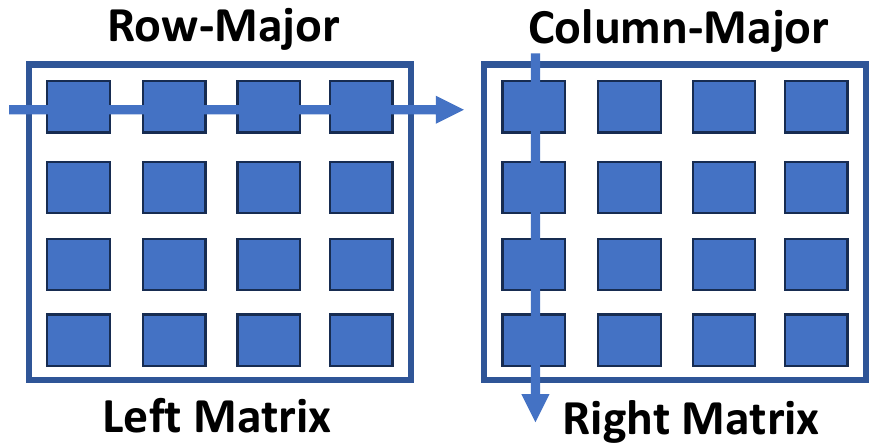}
\caption{Data Layout}
\label{fig:data_layout}
\end{figure}

Figure \ref{fig:data_layout} describes the data layout. In order to compute the matrix product $\bm{A}\bm{B}$, $\bm{A}$ being the left matrix and $\bm{B}$ the right matrix, the left matrix $\bm{A}$ is stored in a row-major format and the right matrix $\bm{B}$ is stored in a column-major format.

The matrices are divided into blocks of size $b \times b$. The matrices are stored in a block-contiguous fashion, either in row-major format (left matrix) or column-major format (right matrix). The elements within each block are stored in a contiguous fashion.

\subsection{Compute Units}
\label{subsubsec: cu_ker}

\begin{figure}[H]
\centering
\includegraphics[width=0.75\linewidth]{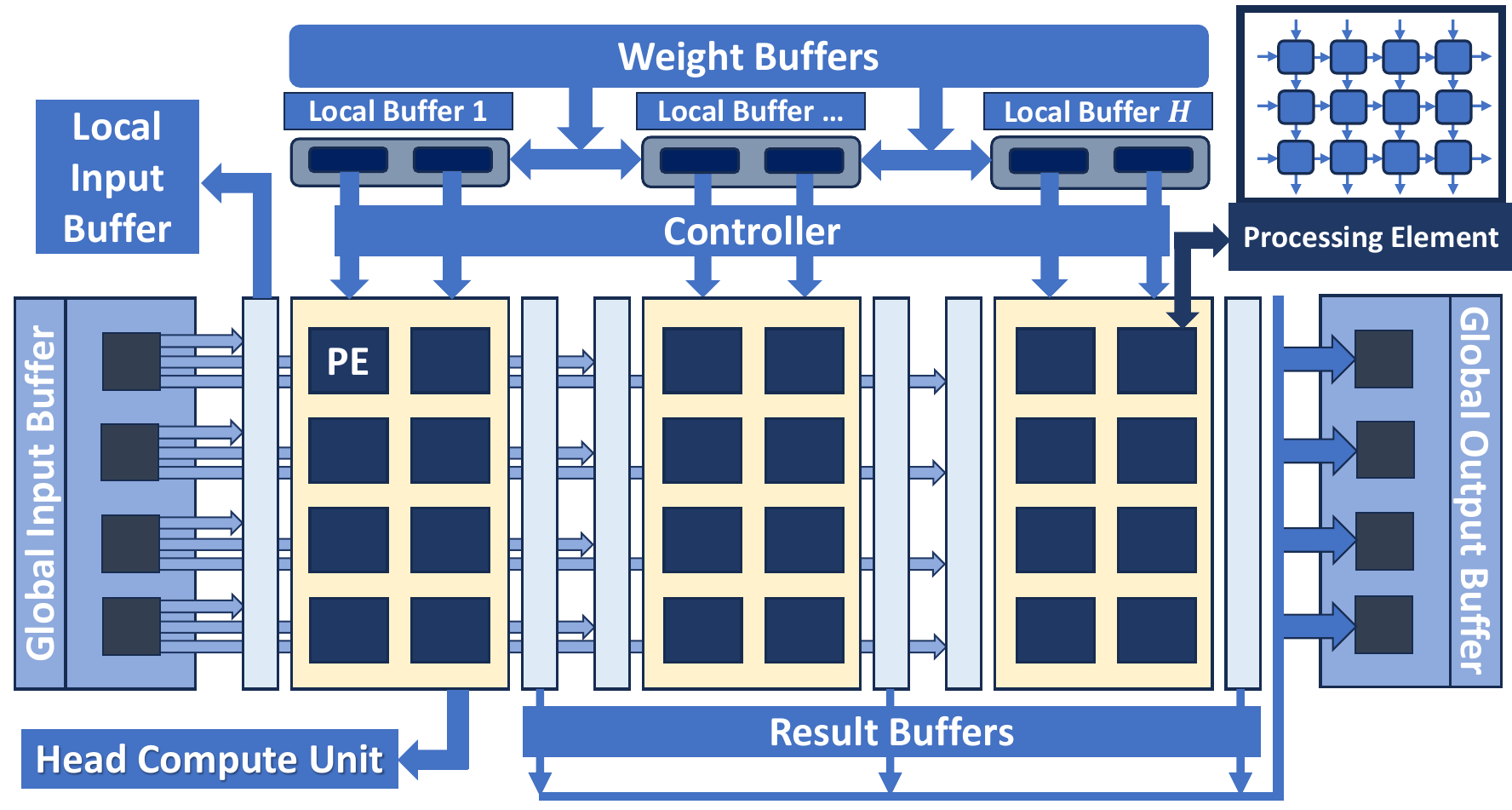}
\caption{Block diagram of the proposed HPPU}
\label{fig:hppu}
\end{figure}

\paragraph{HPPU} 
The core computing unit is show in Figure \ref{fig:hppu}. In order to leverage the inherent compute parallelism along the attention heads within a transformer, the entire compute unit is divided into several Head Compute Units (HCUs). Each HCU computes an individual (attention) head. Further, each HCU contains a 2D mesh of processing elements (PEs). Thus, for a total of $p_h$ HCUs, each HCU comprising of $p_t \times p_c$ PEs, the total number of PEs in the HPPU are $p_h \times p_t \times p_c$. This arrangement of PEs exploits compute parallelism across three levels of compute dimensions. PEs along the $p_t$ and $p_c$ axes exploit compute parallelism within each head along the token and embedding dimensions, respectively. $p_h$ such $p_t \times p_c$ instances of HCUs allows computing $p_h$ heads simultaneously. 

Each PE is organized as a $p_{pe} \times p_{pe}$ grid of compute elements (systolic array) (shown in Figure \ref{fig:hppu}). The Global Input Buffer (GIB) is utilized to store the input feature matrix. Each HCU has its own, individual Local Input Buffer (LIB). Data from the GIB is appropriately streamed (broadcasted) to each HCUs LIB, allowing for simultaneous compute. The output results computed by each HCU are stored in Local Output Buffers (LOB) and are streamed out into a Global Output Buffer (GOB). The input weight matrix is stored in a Weight Buffer (WB). This WB is partitioned into several local buffers (banks) for each column of computing PEs within an HCU.

The HPPU performs Dense Block-wise Matrix Multiplication (DBMM) on two dense matrices. The dense matrices are partitioned block-wise (Section \ref{subsubsec: layout}) into blocks of size $b \times b$. Each PE computes an output block for the output matrix. 

\begin{figure}[ht]
\centering
\includegraphics[scale=0.4]{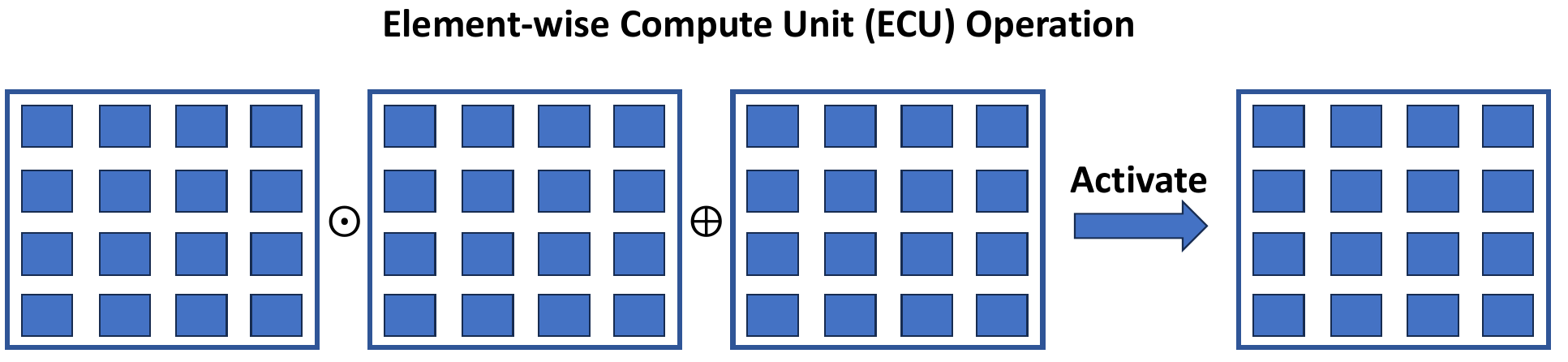}
\caption{Element-wise Compute Unit (ECU) Operation}
\label{fig:ecu}
\end{figure}

\paragraph{ECU} 
The ECU is structured identically to the HPPU (as in Figure \ref{fig:hppu}). However, its main function is to perform element-wise computations (element-wise multiply and/or add) (Figure \ref{fig:ecu}). The ECU can also perform element-wise non-linear activation ($\text{GELU}$) or exponentiation ($\text{exp}$). The ECU has a total of $4$ buffers to support a general operation of the form $f(\bm{A} \bigodot \bm{B} \bigoplus \bm{C})$. Here $f(.)$ represents the non-linear activation, $\bigodot$ respresents element-wise multiply and $\bigoplus$ represents element-wise add. $\bm{A}, \bm{B}$ and $\bm{C}$ are equal sized matrices stored per the layout in Section \ref{subsubsec: layout}.


The $3$ buffers for ECU store the matrices $\bm{A}, \bm{B}$ and $\bm{C}$. A final buffer stores the final computed result. Each PE computes an output block of result through element-wise operations. This is in contrast to computing an output block of result through block-wise matrix multiplication, as in the HPPU.

\subsection{Compute Flow}
\label{subsubsec: com_flow}
In this Section, we describe the overall compute flow for the accelerator in Section \ref{subsubsec: cu_ker} to compute the output for an input $\bm{X}$. The input $\bm{X}$ is shifted and tokenized (Section \ref{subsec:sys_overview}). The compute flow within each VTR encoder layer can be partitioned into two flows: (i) Multi-Headed Self-Attention (MSA) compute (ii) Multi-Layer Perceptron (MLP) compute. Prior to describing the MSA and MLP compute, we first describe the generation of learned embeddings from the shifted tokenized input, $\bm{X}$.

\paragraph{Embedding Generation}
The shifted and tokenized input $\bm{X} \in R^{N \times D'}$, is transformed into learned embeddings as per eq. \ref{eq: embd}. Note that $N$ refers to the total tokens and $D'$ refers to the raw embedding size.

\begin{equation}
    \bm{Z} = \text{concat}(\bm{x_\text{CLS}},\text{Linear}(\text{LN}(\bm{X})))
\label{eq: embd}
\end{equation}

In eq. \ref{eq: embd}, $\bm{x_\text{CLS}}$ refers to the learned class token embedding. $\text{LN}(.)$ stands for layer normalization layer and $\text{Linear}(.)$ stands for a linear layer (a single layer MLP). The learned embeddings $\bm{Z} \in R^{(N + 1) \times D}$ are obtained by performing layer normalization on the input matrix $\bm{X}$ (along the embedding axis). The layer-normed embeddings, $\text{LN}(\bm{X})$, are passed through a linear layer. Finally, the class embedding vector, $\bm{x}_\text{CLS}$, is concatenated to the output of the linear layer to generate the final embeddings, $\bm{Z}$.

The operation of layer normalization is performed via both the HPPU and ECU units. The mean and standard deviation of each embedding vector is computed via ECU. The required aggregations for this are performed by HPPU via a multiply by $\bm{1}$ operation (where $\bm{1}$ refers to a vector of $1$'s). The learned parameters $\bm{\gamma} \, \text{and} \, \bm{\beta} \in R^{D'}$ of the layer normalization layer are finally applied to $\bm{X}$ as $\frac{\bm{X} - \bm{\mu}}{\bm{\sigma}} \bigodot \bm{\gamma} \bigoplus \bm{\beta}$ via the ECU. Here $\bm{\mu} \, \text{and} \, \bm{\sigma} \in R^{D'}$ are the mean and standard deviation vectors for the embeddings $\bm{X}$ (token-wise), respectively.  

The linear layer is mapped to the DBMM operation performed via the HPPU. Note that there is no direct notion of a `head' associated with a simple linear layer. However, the output matrix can be partitioned into several `fictitious heads' along along the column axis. Then, each such `fictitious head' is mapped to an HCU within the HPPU.

\paragraph{MSA Compute}
The MSA compute contains several stages described thus. The generated embedding matrix $\bm{Z}$ is used to generate the $\bm{Q}$, $\bm{K}$ and $\bm{V}$ matrix via weights $\bm{W}_Q$, $\bm{W}_K$ and $\bm{W}_V$ respectively. The $\bm{W}_Q$, $\bm{W}_K$ and $\bm{W}_V$ weights are naturally partitioned per the total heads, $H$, associated with the multi-headed self-attention mechanism within the encoder. Thus, the $\bm{Q}, \bm{K}, \bm{V}$ matrices can be represented as follows,

\begin{equation}
    \bm{Y} = [\bm{Y}_1 \quad \bm{Y}_2 \quad ... \quad \bm{Y}_H] \quad \text{where} \quad \bm{Y} \in \{\bm{Q}, \bm{K}, \bm{V}\} 
\label{eq: head_split}
\end{equation}

In eq. \ref{eq: head_split}, the output matrix $\bm{Y}$ is split into several heads, with the output matrix of some head $i$, $\bm{Y}_i$ computed by an HCU via DBMM. Here, $\bm{Y}$ refers to one of the $\bm{Q}$, $\bm{K}$ and $\bm{V}$ matrices. 

Next, the raw attention matrix for each head is computed as below,

\begin{equation}
    \bm{A} = [\bm{A}_1 \quad \bm{A}_2 \quad ... \quad \bm{A}_H] = \bm{Q}\bm{K}^{T} = [\bm{Q}_1\bm{K}_1^{T} \quad \bm{Q}_2\bm{K}_2^{T} \quad ... \quad \bm{Q}_H\bm{K}_H^{T}]
\label{eq: attnt}
\end{equation}

As is suggested by eq. \ref{eq: attnt}, the attention compute is performed by multiplying the query and key matrices of corresponding heads. Each such output matrix $\bm{A}_i$ (for a head $i$) is computed by an HCU using DBMM. The softmax scores are computed via both the ECU and the HPPU compute units. ECU performs the scaling operations and exponentiation. HPPU performs aggregations to generate the scales required for softmax compute. To incorporate the LSA mechanism, a learned scaling factor $\lambda$ (instead of $\sqrt{d_k}$) is used to scale the raw attention matrix. Further, post-scaling (pre-softmax), the values in the attention matrix, along the diagonals, are set to arbitrarily large negative values. Post-softmax, the attention scores along the diagonal (in each head), thus become close to $0$. This effectively removes a tokens self-value vector in computing the weighted aggregate of value vectors with its attention scores. As a result, the tokens focus more on inter-token relations.

Finally, the processed attention matrix comprising the attention scores are used along with the value matrix as below,
\begin{equation}
    \bm{O} = \bm{S}\bm{V} = [\bm{S}_1 \bm{V}_1 \quad \bm{S}_2 \bm{V}_2 \quad ... \quad \bm{S}_H \bm{V}_H]
\label{eq: AV}
\end{equation} 

In eq. \ref{eq: AV}, $\bm{S}$ (attention score matrix) is computed from $\bm{A}$ as below,
\begin{subequations}\label{eqn_S}
\begin{align*}
    \bm{A}_\lambda &= \bm{A}/\lambda \\
    \bm{A}_\lambda' &= [\text{M}(\bm{A}_{\lambda1}) \quad \text{M}(\bm{A}_{\lambda2}) \quad ... \quad \text{M}(\bm{A}_{\lambda H})] \\
    \bm{S} &= \text{softmax}(\bm{A}_\lambda') \tag{\hypersetup{linkcolor=black}\ref{eqn_S}} 
\end{align*}
\end{subequations}

In eq. \ref{eqn_S}, $\text{M}(.)$ refers to the operation of setting the values along the diagonal of a matrix $\bm{A}_{\lambda i}$ (for the $i^{\text{th}}$ head) to an arbitrarily large value close to $- \infty$.

The compute in eq. \ref{eq: AV} is similar to that in eq. \ref{eq: attnt} and is performed by the HPPU through the DBMM operation. Finally, the projection matrix $\bm{W}_p$ is utilized to compute the final MSA output as $\text{MSA}(\bm{X})$ = $\bm{O}$$\bm{W}_p$. This compute is similar to computing the $\bm{Q}, \bm{K}, \bm{V}$ matrices and is also performed via the HPPU through the DBMM operation.

\paragraph{MLP Compute} The compute in the MLP stage of the encoder comprises of two single layer feed-forward neural networks. This is performed as DBMM via the HPPU similar to the description in the MSA compute stage. 

\section{EVALUATION}
\subsection{Experimental Setting}
 We implement the VTR model using PyTorch 2.0.1\cite{NEURIPS2019_bdbca288} and utilize an NVIDIA RTX A6000 GPU with CUDA 11.8 for training the model. Several model variants are trained to comprehensively evaluate the VTR model's robustness and generalization across different small-sized SAR datasets. Specifically, we vary the following hyper-parameters:
\begin{itemize}
 \item \textbf{Patch Size:} We explore different patch sizes for partitioning the input image, to assess its impact on model performance.
 \item \textbf{Hidden Dimension:} The dimensionality of the hidden layers within the transformer encoder is varied to observe its effects on model learning.
 \item \textbf{Depth:} In our experiments, we explore varying depths of layers in the transformer encoder to assess the model's sensitivity to depth.
 \item \textbf{Number of Heads:} We use two settings of attention heads in the multi-headed self-attention mechanism to analyze its influence on model behavior.
\end{itemize}
We directly train the VTR model on three distinct SAR datasets: MSTAR \cite{MSTAR}, SynthWakeSAR \cite{rizaev2022synthwakesar_synthwakesar}, and GBSAR \cite{10133345_gbsar}. The three datasets represent a diverse range of SAR imaging scenarios and characteristics. The Adam optimization algorithm \cite{kingma2014adam} is utilized for training, with an initial learning rate of $0.001$. Subsequently, the step learning rate optimizer is applied, with a step size of $20$ and a gamma value of $0.5$. The model is trained for $200$ epochs.

\subsection{Hardware Implementation Details}
The HPPU and ECU described in Section \ref{subsec: hw_des} are implemented on a state-of-the-art FPGA platform, Xilinx Alveo U250. The main hyper-parameters associated with the proposed accelerator, $p_h, p_t, p_c$ and $p_{pe}$ are selected as $4, 12, 2$ and $8$, respectively. $p_h$ of $4$ allows the accelerator to fully utilize the $4$ SLRs (Super Logic Regions) on the FPGA. $p_t$ was selected based on the nominal input token blocks (associated with the input image). $p_c$ of $2$ allows for both the PE columns within each HCU (in each SLR) to concurrently access the FPGA BRAM/URAM using its dual port memory. $p_{pe}$ of $8$ is selected to support any block size $b$ which is a multiple of $8$ (such as $8, 16, 32$, etc.). The accelerator is designed using Xilinx High Level Synthesis (HLS) and synthesized using Xilinx Vitis v2022.2 with an achieved frequency of $300$ MHz. 

\subsection{Datasets}
\textbf{MSTAR:} The original Moving and Stationary Target Acquisition and Recognition (MSTAR) dataset consists of 5172 
 SAR images. It contains 10 categories of ground vehicles, with 2747 images in the training set and 2425 images in the testing set. Each image is of $128\times 128$ pixels. For our experiments, we use an augmented MSTAR dataset which consists 27,000 SAR images in the training set and 2425 images in the testing set. The size of each image in this dataset is $88\times 88$. Hence, the images in this dataset are generated by cropping the images in the original MSTAR dataset in various directions. \\
\textbf{SynthWakeSAR:} SynthwakeSAR is a synthetic SAR imagery dataset comprising 92,160 images of 10 different real vessel models. There are 73,728 images in the training set and 18,432 images in the test set. Each image is of $128\times 128$ pixels. The images of each vessel contain the ship's wakes. \\
\textbf{GBSAR:} Ground Based SAR (GBSAR) dataset consists of 6434 raw radar images of which 5147 images are in the training set and 1287 images are in the test set. This dataset captures 7 different ceramic cups with rubber objects. Each raw SAR image has a size of $88\times 88$.

\subsection{Platform Specification} We evaluate the performance of our accelerator for the SAR ATR applications against state-of-the-art CPU and GPU platforms, with the specifications of the platforms detailed in Table \ref{tab:plat}. Table \ref{tab:plat} also contains specifications associated with the FPGA platform utilized for the hardware accelerator.

\begin{table}[h!]
\centering
\caption{Specifications of platforms}
\label{tab:plat}
\begin{tabular}{cccc} 
\toprule
& \textbf{CPU} & \textbf{GPU} &  \textbf{FPGA} \\ 
\midrule
\midrule
\textbf{Platform} &  \begin{tabular}[|c|]{@{}c@{}} AMD \\ EPYC 9654 \end{tabular} & \begin{tabular}[|c|]{@{}c@{}}  NVIDIA RTX \\ 6000 Ada \end{tabular}   & \begin{tabular}[|c|]{@{}c@{}} Xilinx \\ Alveo U250 \end{tabular} \\  \midrule
\textbf{Frequency} & 2.4 GHz & 915 MHz & 300 MHz \\ \midrule  
\begin{tabular}[|c|]{@{}c@{}} \textbf{Peak} \\ \textbf{Performance} \\  \textbf{(TFLOPS)} \end{tabular}& 3.69 & 91.06  & 1.8 \\ \midrule  
\begin{tabular}[|c|]{@{}c@{}}  \textbf{On-chip} \\ \textbf{Memory}  \end{tabular} & 384 MB & 96MB & 36 MB \\ \midrule
\begin{tabular}[|c|]{@{}c@{}} \textbf{Memory} \\ \textbf{Bandwidth}  \end{tabular}  & 461 GB/s & 960 GB/s & 77 GB/s\\
\bottomrule
\end{tabular}
\end{table}

\subsection{Performance Metrics}
\label{subsec:perf}
We evaluate our model using the following metrics: classification accuracy, total MACs (multiply and accumulate operations), model parameters, inference latency and throughput. Classification accuracy is measured as the ratio of images correctly classified by the model. MACs and model parameters characterize the computational complexity associated with a given model; MACs measures the total multiply-accumulate operations that a model performs to compute the output, model parameters are the total learnable parameters in a model. The inference latency is defined as the end-to-end latency (time) required for a model to compute the output given an input image, for a given platform. Finally, throughput is defined as the total images that can be processed on a given platform in a second, where processed refers to computing the output for an input image.

\subsection{Experimental Results}
The experimental results obtained across all three datasets are summarized in Tables \ref{tab:mstar}-\ref{tab:gbsar}. Considering that the images in the MSTAR and GBSAR datasets are sized $88\times 88$, our experiments involve patch sizes of 8 and 11, along with hidden dimension sizes of 44 and 88, respectively. However, as the image size in the SynthwakeSAR dataset is $128\times 128$, we opt for patch sizes of 8 and 16, accompanied by hidden dimension sizes of 96 and 48, respectively. The depth of the transformer encoder is set to 4, 6, 8, or 12, and the number of heads is configured to be 2 or 4. Each of these hyperparameter settings is employed during the training of the model to assess its performance across different configurations. Latency for FPGA, GPU and CPU platforms is the end-to-end latency defined in \ref{subsec:perf}, measured from the time that the input is sent to a processor to the time that the processor takes to compute the output. Throughput for the three platforms is defined as per \ref{subsec:perf} measuring the total images that can be processed in a given second. For latency comparison across the $3$ platforms, we utilize a batch size of $1$. A batch size of $1$ represents a real-time scenario for SAR ATR applications wherein the input images are continuously streaming. In order to compare the throughput for the $3$ platforms, results are computed over batch sizes $1, 8, 16, 32, 64, 128, 256$ and $512$. Note that the FPGA throughput is nominally measured as the inverse of the per-image inference latency, and thus is invariant with respect to batch size.
\begin{table}[hbt]
\centering
\caption{Performance on MSTAR dataset}
\vspace{3pt}
    \label{tab:mstar}
    \begin{adjustbox}{max width=0.8\textwidth}
\begin{tabular}{cccccccccc}
\hline
\multirow{2}{*}{Patch size} & \multirow{2}{*}{Hidden dimension} & \multirow{2}{*}{Depth} & \multirow{2}{*}{\# Heads} & \multirow{2}{*}{Accuracy} & \multirow{2}{*}{\# MACs} & \multirow{2}{*}{\# Parameters} & \multicolumn{3}{c}{Latency (ms)} \\
 & & & & & & & CPU & GPU  & FPGA  \\
 \hline
\multirow{16}{*}{8} & \multirow{8}{*}{44}& \multirow{2}{*}{4}  & 2   & 89.61\%  & 0.878G  & 109.99K  &  17.64 & 2.03    &  0.037 \\\cline{4-10}
&                    &                     & 4   & 87.84\%  & 0.901G  & 109.99K  & 10.05  &  2.16   &  0.034 \\\cline{3-10} 
&                    & \multirow{2}{*}{6}  & 2   & 90.93\%  & 1.26G    & 157.33K  & 10.99  &  2.95  & 0.056
\\\cline{4-10}
&                    &                     & 4   & 89.57\%  & 1.29G    & 157.33K  &  11.10 &  2.97   &  0.052\\\cline{3-10} 
&                    & \multirow{2}{*}{8}  & 2   & 91.05\%  & 1.64G    & 204.68K  & 11.35  &  3.84   & 0.075  \\\cline{4-10}
&                    &                     & 4   & 90.68\%  & 1.68G    & 204.68K  & 13.24  & 3.85    &  0.069 \\\cline{3-10} 
&                    & \multirow{2}{*}{12} & 2   & 94.06\%  & 2.40G    & 299.37K  & 13.17  & 5.61    &  0.11 \\\cline{4-10}
&                    &                     & 4   & 91.51\%  & 2.47G    & 299.37K  & 13.62  & 5.62    &  0.10 \\\cline{2-10} \cline{2-10}\cline{2-10}
& \multirow{8}{*}{88}& \multirow{2}{*}{4}  & 2   & 90.47\%  & 3.18G    & 405.19K  & 5.04  & 2.18    &  0.067\\\cline{4-10}
&                    &                     & 4   & 91.30\%  & 3.20G    & 405.19K  & 5.21  &  2.18   & 0.064 \\\cline{3-10} 
&                    & \multirow{2}{*}{6}  & 2   & 93.98\%  & 4.65G    & 592.80K  & 7.15  & 2.98    & 0.10 \\\cline{4-10}
&                    &                     & 4   & 93.94\%  & 4.68G    & 592.80K  & 7.39  & 2.99    &  0.097  \\\cline{3-10} 
&                    & \multirow{2}{*}{8}  & 2   & 94.35\%  & 6.12G    & 780.42K  & 9.30  & 3.88    &  0.13 \\\cline{4-10}
&                    &                     & 4   & 93.86\%  & 6.17G    & 780.42K  & 9.75  & 3.89    &  0.13 \\\cline{3-10} 
&                    & \multirow{2}{*}{12} & 2   & 95.18\%  & 9.07G    & 1156K  & 13.80  & 5.71    & 0.20  \\\cline{4-10}
&                    &                     & 4   & \textbf{95.96\%}  & \textbf{9.14G}    & \textbf{1156K}  & \textbf{14.41}  & \textbf{5.71}    &  \textbf{0.19} \\
\hline
\cline{1-10}\cline{1-10}
\multirow{16}{*}{11} & \multirow{8}{*}{44}& \multirow{2}{*}{4}  & 2   & 86.27\%  & 0.517G  & 123.10K  & 3.50  & 2.17   & 0.029  \\\cline{4-10}
&                    &                     & 4   & 84.33\%  & 0.524G & 123.10K & 3.61  & 2.17  & 0.028  \\\cline{3-10} 
&                    & \multirow{2}{*}{6}  & 2   & 88.25\%  & 0.717G & 170.44K & 5.05  & 2.94  & 0.043  \\\cline{4-10}
&                    &                     & 4   & 89.49\%  & 0.727G & 170.44K & 5.18  & 2.95  &  0.042 \\\cline{3-10} 
&                    & \multirow{2}{*}{8}  & 2   & 90.56\%  & 0.917G & 217.79K & 6.55  & 3.84  & 0.058  \\\cline{4-10}
&                    &                     & 4   & 89.81\%  & 0.930G & 217.79K & 6.60  & 3.83  &  0.056  \\\cline{3-10} 
&                    & \multirow{2}{*}{12} & 2   & 91.96\%  & 1.32G   & 312.48K & 9.53  & 5.61  & 0.087  \\\cline{4-10}
&                    &                     & 4   & 91.05\%  & 1.34G   & 312.48K & 9.67  & 5.62  & 0.084  \\\cline{2-10} \cline{2-10}\cline{2-10}
& \multirow{8}{*}{88}& \multirow{2}{*}{4}  & 2   & 90.23\%  & 1.79G   & 430.83K & 4.02  & 2.16  &  0.052 \\\cline{4-10}
&                    &                     & 4   & 89.49\%  & 1.80G   & 430.83K & 4.10  & 2.17  & 0.051 \\\cline{3-10} 
&                    & \multirow{2}{*}{6}  & 2   & 91.92\%  & 2.58G   & 618.45K & 5.79  & 2.95  & 0.078 \\\cline{4-10}
&                    &                     & 4   & 91.75\%  & 2.58G   & 618.45K & 5.87  & 2.95  & 0.076  \\\cline{3-10} 
&                    & \multirow{2}{*}{8}  & 2   & 93.94\%  & 3.36G   & 806.07K & 7.59  & 3.85  & 0.10  \\\cline{4-10}
&                    &                     & 4   & 92.62\%  & 3.37G   & 806.07K & 7.62  & 3.85  & 0.10  \\\cline{3-10} 
&                    & \multirow{2}{*}{12} & 2   & 94.72\%  & 4.92G   & 1181K & 10.97  & 5.63    & 0.15  \\\cline{4-10}
&                    &                     & 4   & 93.24\%  & 4.94G   & 1181K & 11.29  & 5.66   & 0.15 \\
\hline
\cline{1-10}\cline{1-10}
\end{tabular}
\end{adjustbox}
\end{table}

\begin{table}[hbt]
\centering
\caption{Performance on SynthWakeSAR dataset}
\vspace{3pt}
    \label{tab:synthwake}
    \begin{adjustbox}{max width=0.8\textwidth}
\begin{tabular}{cccccccccc}
\hline
\multirow{2}{*}{Patch size} & \multirow{2}{*}{Hidden dimension} & \multirow{2}{*}{Depth} & \multirow{2}{*}{\# Heads} & \multirow{2}{*}{Accuracy} & \multirow{2}{*}{\# MACs} & \multirow{2}{*}{\# Parameters} & \multicolumn{3}{c}{Latency (ms)} \\
 & & & & & & & CPU & GPU  & FPGA  \\
 \hline
\multirow{16}{*}{8} & \multirow{8}{*}{48}& \multirow{2}{*}{4}  & 2   & 90.04\%  & 2.22G   & 129.15K  & 5.43  & 2.20    & 0.085  \\\cline{4-10}
&                    &                     & 4   & 90.57\%  & 2.32G  & 129.15K  &  5.39  & 2.21    & 0.062  \\\cline{3-10} 
&                    & \multirow{2}{*}{6}  & 2   & 91.02\%  & 3.19G  & 185.40K  & 7.67  & 3.01    & 0.128  \\\cline{4-10}
&                    &                     & 4   & 91.16\%  & 3.34G  & 185.40K  & 8.17  & 3.01    & 0.093  \\\cline{3-10} 
&                    & \multirow{2}{*}{8}  & 2   & 91.33\%  & 4.16G  & 241.66K  & 10.15  & 3.90    & 0.17  \\\cline{4-10}
&                    &                     & 4   & 91.57\%  & 4.37G  & 241.66K  & 10.90  & 3.92    &  0.12  \\\cline{3-10} 
&                    & \multirow{2}{*}{12} & 2   & 92.16\%  & 6.11G  & 354.17K  & 14.83  & 5.72    & 0.25  \\\cline{4-10}
&                    &                     & 4   & 92.18\%  & 6.41G  & 354.17K  & 15.78  & 5.72    & 0.18  \\\cline{2-10} \cline{2-10}\cline{2-10}
& \multirow{8}{*}{96}& \multirow{2}{*}{4}  & 2   & 92.55\%  & 7.95G  & 478.83K  & 6.82  & 2.23    & 0.17  \\\cline{4-10}
&                    &                     & 4   & 92.56\%  & 8.06G  & 478.83K  & 6.64  & 2.23    & 0.16  \\\cline{3-10} 
&                    & \multirow{2}{*}{6}  & 2   & 93.34\%  & 11.67G  & 701.93K  & 9.70  & 3.05    & 0.25  \\\cline{4-10}
&                    &                     & 4   & 93.35\%  & 11.82G  & 701.93K  & 9.43  & 3.05    & 0.24  \\\cline{3-10} 
&                    & \multirow{2}{*}{8}  & 2   & 93.05\%  & 15.38G  & 925.03K  & 12.76  & 3.98    & 0.34  \\\cline{4-10}
&                    &                     & 4   & 93.44\%  & 15.58G  & 925.03K  &  12.17  & 3.97    & 0.32  \\\cline{3-10} 
&                    & \multirow{2}{*}{12} & 2   & 93.21\%  & 22.80G  & 1371K  &  18.39 & 5.82    &  0.51 \\\cline{4-10}
&                    &                     & 4   & \textbf{93.47\%}  & \textbf{23.11G}  & \textbf{1371K}  & \textbf{17.83}   &  \textbf{5.81}   & \textbf{0.48}  \\
\hline
\cline{1-10}\cline{1-10}
\multirow{16}{*}{16} & \multirow{8}{*}{48}& \multirow{2}{*}{4}  & 2   & 84.83\%  & 0.745G  & 177.15K  & 3.61  & 2.19  & 0.029  \\\cline{4-10}
&                    &                     & 4   & 85.35\%  & 0.752G  & 177.15K  & 3.72  & 2.07    & 0.029  \\\cline{3-10} 
&                    & \multirow{2}{*}{6}  & 2   & 86.52\%  & 0.982G  & 233.40K  & 5.15  & 2.98    & 0.044 \\\cline{4-10}
&                    &                     & 4   & 87.23\%  & 0.992G  & 233.40K  & 5.31  &  2.98   & 0.043  \\\cline{3-10} 
&                    & \multirow{2}{*}{8}  & 2   & 87.66\%  & 1.22G    & 289.66K  & 6.69  & 3.88    &  0.059 \\\cline{4-10}
&                    &                     & 4   & 88.01\%  & 1.23G    & 289.66K  & 6.85  & 3.88    & 0.058  \\\cline{3-10} 
&                    & \multirow{2}{*}{12} & 2   & 89.45\%  & 1.69G    & 402.17K  & 9.57  & 5.67    & 0.089  \\\cline{4-10}
&                    &                     & 4   & 88.53\%  & 1.71G    & 402.17K  & 9.81  & 5.67    & 0.087  \\\cline{2-10} \cline{2-10}\cline{2-10}
& \multirow{8}{*}{96}& \multirow{2}{*}{4}  & 2   & 90.29\%  & 2.38G    & 572.91K  & 4.30  & 2.19    & 0.053  \\\cline{4-10}
&                    &                     & 4   & 89.99\%  & 2.39G    & 572.91K  & 4.29  & 2.20    & 0.052 \\\cline{3-10} 
&                    & \multirow{2}{*}{6}  & 2   & 90.75\%  & 3.31G    & 796.01K  & 6.11  &  3.01   & 0.08  \\\cline{4-10}
&                    &                     & 4   & 91.22\%  & 3.32G    & 796.01K  & 6.09  & 3.02    & 0.079  \\\cline{3-10} 
&                    & \multirow{2}{*}{8}  & 2   & 91.95\%  & 4.24G    & 1019K  & 7.80  & 3.94    &  0.10 \\\cline{4-10}
&                    &                     & 4   & 91.25\%  & 4.26G    & 1019K  & 7.83  & 3.93    & 0.10  \\\cline{3-10} 
&                    & \multirow{2}{*}{12} & 2   & 91.91\%  & 6.10G    & 1465K  & 11.26  & 5.74    & 0.16  \\\cline{4-10}
&                    &                     & 4   & 91.85\%  & 6.12G    & 1465K  & 11.49  &  5.76   & 0.15 \\
\hline
\cline{1-10}\cline{1-10}
\end{tabular}
\end{adjustbox}
\end{table}

\begin{table}[hbt]
\centering
\caption{Performance on GBSAR dataset}
\vspace{3pt}
\label{tab:gbsar}
\begin{adjustbox}{max width=0.8\textwidth}
\begin{tabular}{cccccccccc}
\hline
\multirow{2}{*}{Patch size} & \multirow{2}{*}{Hidden dimension} & \multirow{2}{*}{Depth} & \multirow{2}{*}{\# Heads} & \multirow{2}{*}{Accuracy} & \multirow{2}{*}{\# MACs} & \multirow{2}{*}{\# Parameters} & \multicolumn{3}{c}{Latency (ms)} \\
 & & & & & & & CPU & GPU  & FPGA  \\
 \hline
\multirow{16}{*}{8} & \multirow{8}{*}{44}& \multirow{2}{*}{4}  & 2   & 98.99\%  & 0.878G  & 109.86K  & 4.44  & 2.06    & 0.037  \\\cline{4-10}
&                    &                     & 4  & 98.45\%  & 0.901G  & 109.86K  & 4.60  &  2.18   &  0.034 \\\cline{3-10} 
&                    & \multirow{2}{*}{6}  & 2  & 98.76\%  & 1.26G    & 157.20K  & 6.24  &   2.97  &  0.056 \\\cline{4-10}
&                    &                     & 4  & 98.91\%  & 1.29G    & 157.20K  & 6.61  & 2.98   &  0.052 \\\cline{3-10} 
&                    & \multirow{2}{*}{8}  & 2  & 99.07\%  & 1.64G    & 204.54K  & 8.06  &  3.86  &  0.075 \\\cline{4-10}
&                    &                     & 4  & 98.91\%  & 1.68G    & 204.54K  & 8.54  &  3.88   & 0.069  \\\cline{3-10} 
&                    & \multirow{2}{*}{12} & 2  & 98.91\%  & 2.40G    & 299.23K  & 11.65  &  5.65  & 0.11  \\\cline{4-10}
&                    &                     & 4  & 99.07\%  & 2.47G    & 299.23K  &  12.72 &  5.66   &  0.10 \\\cline{2-10} \cline{2-10}\cline{2-10}
& \multirow{8}{*}{88}& \multirow{2}{*}{4}  & 2  & 99.30\%  & 3.18G    & 404.92K  &  5.19 &  2.19   & 0.067  \\\cline{4-10}
&                    &                     & 4  & 99.30\%  & 3.20G    & 404.92K  & 5.39  &  2.20  & 0.064 \\\cline{3-10} 
&                    & \multirow{2}{*}{6}  & 2  & \textbf{99.46\%}  & \textbf{4.65G}    & \textbf{592.54K}  &  \textbf{7.26} & \textbf{3.01}   &  \textbf{0.10} \\\cline{4-10}
&                    &                     & 4  & 98.99\%  & 4.68G    & 592.54K  & 7.78  & 3.02    &  0.097 \\\cline{3-10} 
&                    & \multirow{2}{*}{8}  & 2  & 99.07\%  & 6.12G    & 780.15K  & 9.52  &  3.92   &  0.13 \\\cline{4-10}
&                    &                     & 4  & 98.76\%  & 6.17G    & 780.15K  & 10.03  &  3.92   & 0.13 \\\cline{3-10} 
&                    & \multirow{2}{*}{12} & 2  & 99.30\%  & 9.07G    & 1155K  & 13.84  &  5.74   & 0.20  \\\cline{4-10}
&                    &                     & 4  & 98.91\%  & 9.14G    & 1155K  & 14.38  & 5.75    & 0.19 \\
\hline
\cline{1-10}\cline{1-10}
\multirow{16}{*}{11} & \multirow{8}{*}{44}& \multirow{2}{*}{4}  & 2   & 97.75\%  & 0.518G  & 122.97K  & 3.54  & 2.18  & 0.029  \\\cline{4-10}
&                    &                     & 4  & 98.37\%  & 0.524G  & 122.97K  & 3.64  & 2.18    & 0.028  \\\cline{3-10} 
&                    & \multirow{2}{*}{6}  & 2  & 98.37\%  & 0.717G  & 170.31K  & 5.10  & 2.97   & 0.043  \\\cline{4-10}
&                    &                     & 4  & 98.21\%  & 0.727G  & 170.31K  & 5.37  & 2.97    & 0.042  \\\cline{3-10} 
&                    & \multirow{2}{*}{8}  & 2  & 98.60\%  & 0.917G  & 217.65K  & 6.61  & 3.87    & 0.058  \\\cline{4-10}
&                    &                     & 4  & 98.76\%  & 0.930G  & 217.65K  & 6.81  &   3.88  & 0.056  \\\cline{3-10} 
&                    & \multirow{2}{*}{12} & 2  & 98.37\%  & 1.32G    & 312.34K  & 9.52  & 5.66   &  0.087 \\\cline{4-10}
&                    &                     & 4  & 97.90\%  & 1.34G    & 312.34K  & 9.85  & 5.66    & 0.084  \\\cline{2-10} \cline{2-10}\cline{2-10}
& \multirow{8}{*}{88}& \multirow{2}{*}{4}  & 2  & 98.29\%  & 1.79G    & 430.57K  & 4.16  & 2.17    & 0.052  \\\cline{4-10}
&                    &                     & 4  & 98.60\%  & 1.80G    & 430.57K  & 4.21  & 2.18    & 0.051  \\\cline{3-10} 
&                    & \multirow{2}{*}{6}  & 2  & 99.15\%  & 2.58G    & 618.19K  & 5.92  &  2.98   & 0.078  \\\cline{4-10}
&                    &                     & 4  & 99.07\%  & 2.58G    & 618.19K  & 5.99  &  2.99   & 0.076  \\\cline{3-10} 
&                    & \multirow{2}{*}{8}  & 2  & 99.22\%  & 3.36G    & 805.80K  &  7.62  &  3.86   & 0.10  \\\cline{4-10}
&                    &                     & 4  & 98.68\%  & 3.37G    & 805.80K  &  7.76 &  3.88   & 0.10  \\\cline{3-10} 
&                    & \multirow{2}{*}{12} & 2  & 99.07\%  & 4.92G    & 1181K & 11.11 & 5.67    & 0.15  \\\cline{4-10}
&                    &                     & 4  & 98.45\%  & 4.94G    & 1181K & 11.20  & 5.68    & 0.15 \\
\hline
\cline{1-10}\cline{1-10}
\end{tabular}
\end{adjustbox}
\end{table}

\subsection{Discussion}
Through the experiments, we observe that the performance of the VTR model varies across the three datasets for different hyperparameter settings. Notably, the best classification accuracies (highlighted in bold in Tables \ref{tab:mstar}-\ref{tab:gbsar}) for the MSTAR and SynthWakeSAR datasets (95.96\% and 93.47\% respectively) are achieved with similar hyperparameter configurations. However, for the GBSAR dataset, a lower depth size results in better performance. It can be concluded that a lower patch size results in a higher classification accuracy as smaller patch sizes are capable of capturing fine-grained details in the input SAR images. It is also worth noting that the SPT and LSA modules are crucial for the performance improvements achieved by the VTR model, as the standard ViT performs poorly on SAR datasets unless pre-trained on significantly larger datasets.

In comparison to the benchmarking conducted by Ye et al.\cite{fein2023benchmarking} on the three datasets using CNNs, GNNs, and ViTs, our VTR model equipped with the SPT and LSA modules demonstrates improved or similar performance. Particularly, our model outperforms the state-of-the-art for the SynthWakeSAR dataset, achieving an accuracy of 93.47\%. For the GBSAR dataset, we achieve a comparable classification accuracy of 99.46\%. For the MSTAR dataset, however, GNNs outperform all other models. This is attributed to the fact that images in this dataset contain the actual target ground vehicle object to be classified only within a few pixels in the center of the image. This limits VTRs capability to extract localized features from mostly non-informant tokens, despite the addition of the SPT and LSA modules. In contrast, GNNs excel at capturing this local information, leading to a better performance. These comparison results are summarized in Table \ref{tab:result comparison}. We define the best-performing model as the model that achieves the highest classification accuracy. Table \ref{tab:result comparison 2} compares the number of parameters in the best-performing models. In contrast to previous work, our model has fewer parameters, leading to a smaller model size. Compared against standard pre-trained ViTs with total parameters of the order $10^{8}$, VTR is significantly smaller ($\times100$) with the largest model size being of the order $10^{6}$ parameters.

\begin{figure}[h]
\centering
\begin{minipage}{0.3\textwidth}
  \centering
  \includegraphics[scale=0.5]{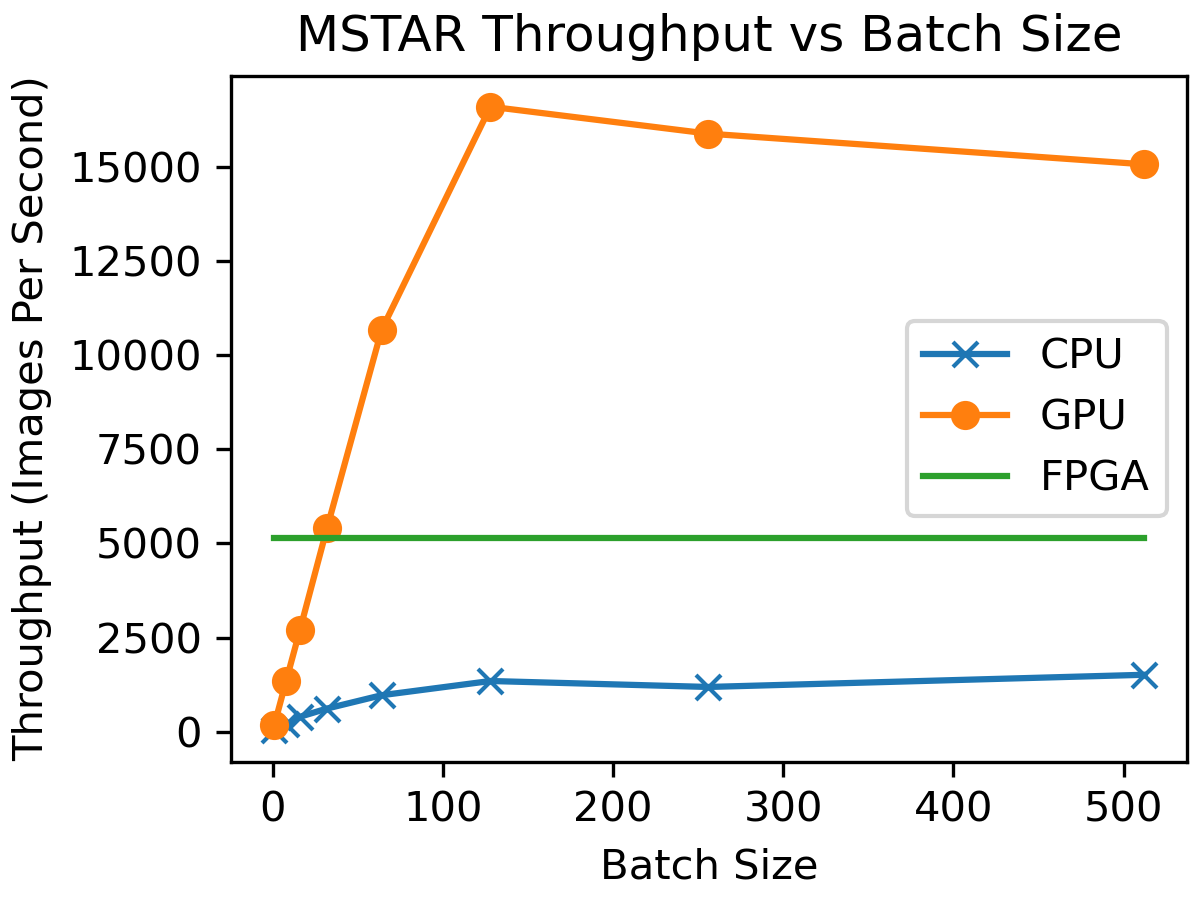}
\end{minipage}
\begin{minipage}{0.3\textwidth}
  \centering
    \includegraphics[scale=0.5]{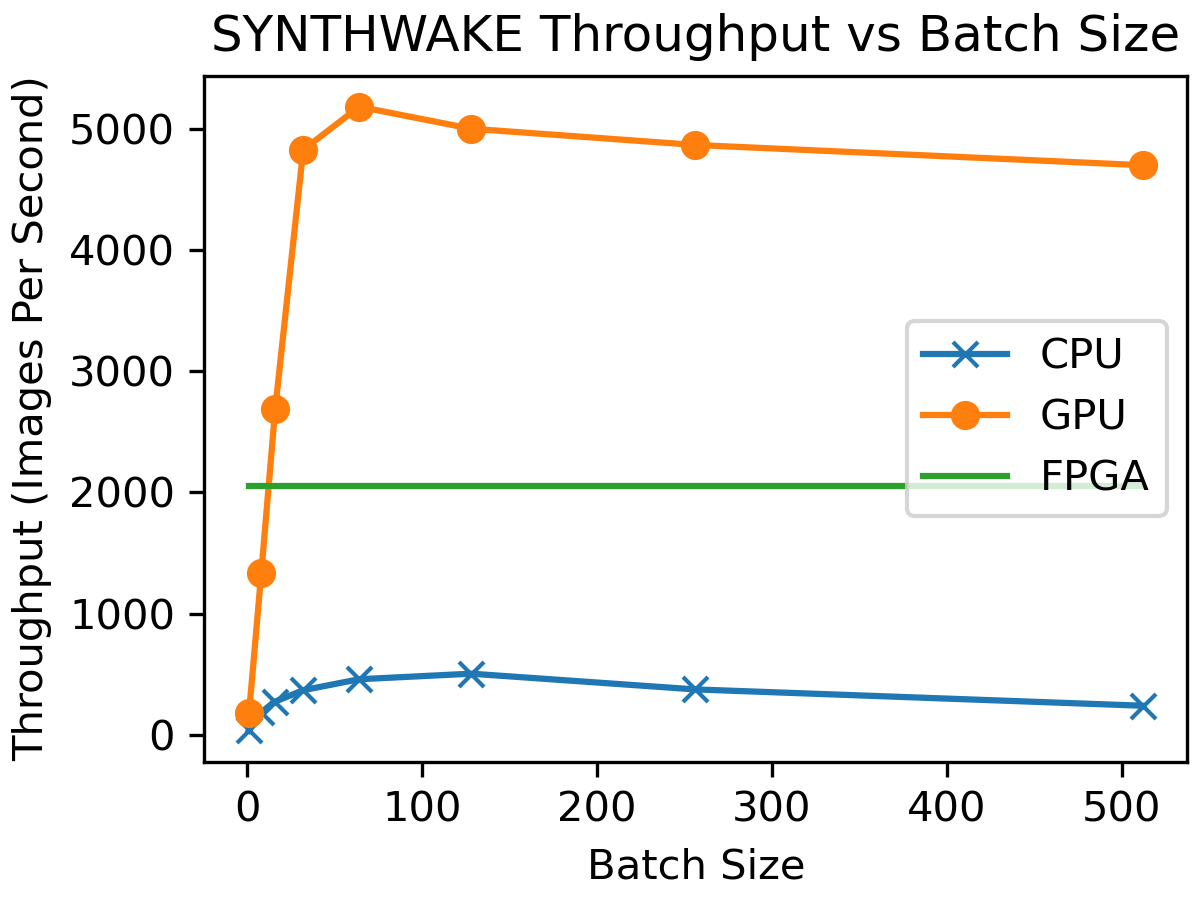}
\end{minipage}
\begin{minipage}{0.3\textwidth}
  \centering
    \includegraphics[scale=0.5]{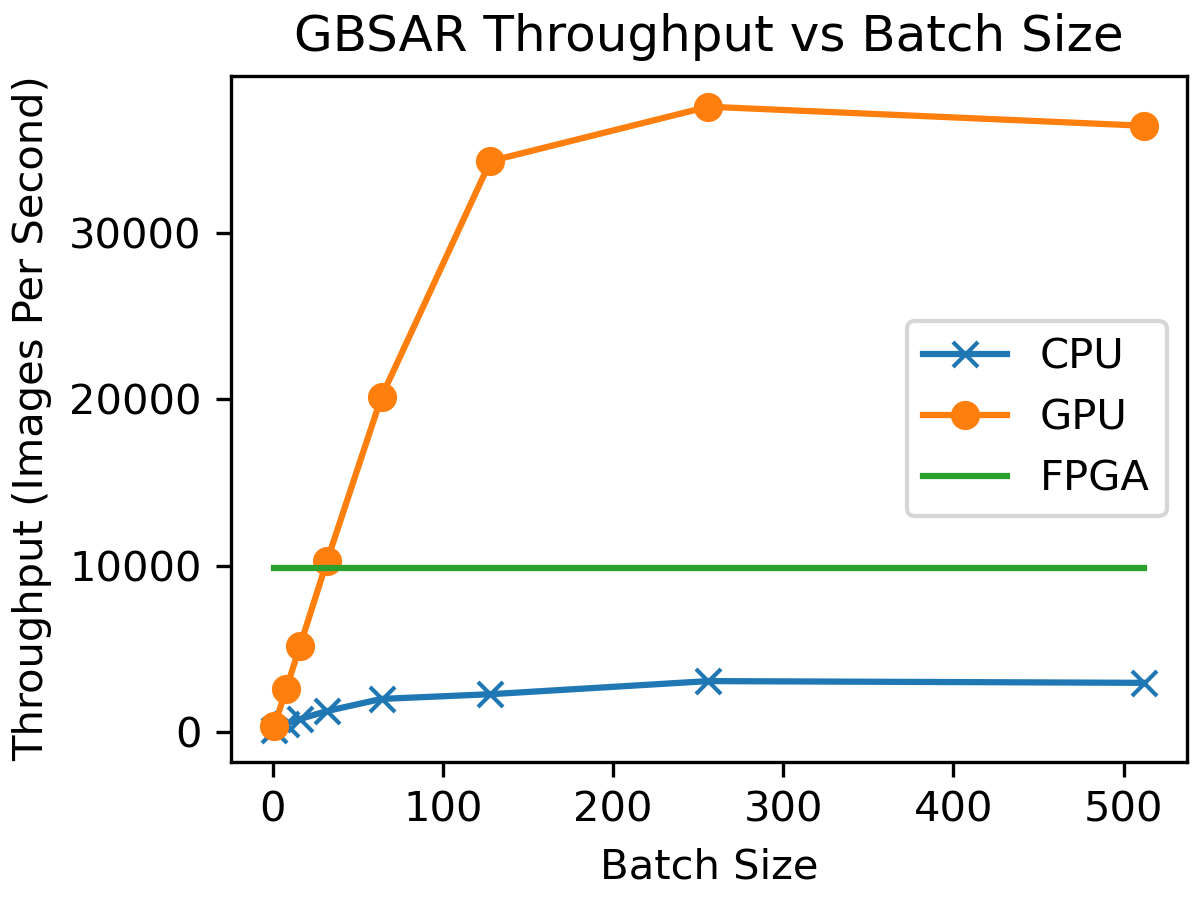}
\end{minipage}
\caption{Throughput on CPU, GPU, and FPGA}
\label{fig:thru}
\end{figure}

The proposed FPGA accelerator, for the best-performing model across the three datasets, has an average speedup (based on single image inference latency) of $\times30$ and $\times70$ when compared to GPU and CPU platforms, respectively. For smaller models, the accelerator has a much higher speedup than the nominal speedup (associated with larger models). Thus, it is highly suitable for deployment in real-time SAR ATR workloads. Figure \ref{fig:thru} compares the throughput of the best-performing model (for each dataset) versus the baseline FPGA throughput (green line). For smaller batch sizes of $1, 8, 16$ and $32$, the FPGA throughput is much better compared to its CPU and GPU counterparts. With larger batch sizes, as expected, the GPU throughput overtakes the FPGA throughput. This suggests that despite being optimized for streaming single image input inferences, our proposed FPGA accelerator can also be used for smaller batch sizes such as $8$ or $16$ with a throughput that is still, on average, $\times2$ that of the GPU throughput. 


\begin{table}[H]
\centering
\caption{Comparison of classification accuracy}
    \label{tab:result comparison}
\begin{tabular}{ccccccc}
\hline
\multirow{2}{*}{Dataset} & \multicolumn{6}{c}{Model}  \\ \cline{2-7} 
              & VTR       & ResNet18\cite{he2016deep} & ResNet34\cite{he2016deep}  & ResNet50\cite{he2016deep} & SS-ViT\cite{lee2021vision}  & Multi-layer GNN\cite{10035150_bingyi_sar_atr}   \\ \hline
MSTAR         & 95.96\%    & 98.47\%  & 98.64\%  & 90.34\%  & 95.61\% & \textbf{99.09\%} \\
SynthWakeSAR  & \textbf{93.47\%} & 90.30\%  & 92.14\%  & 92.42\%  & 87.98\% & 91.15\%          \\
GBSAR         & 99.46\%     & 99.30\%  & \textbf{99.99\%} & 99.53\%  & 99.04\% & 98.67\% \\ \hline        
\end{tabular}
\end{table}

\begin{table}[H]
\centering
\caption{Comparison of the \# parameters in the best performing models with prior work}
\vspace{3pt}
    \label{tab:result comparison 2}
\begin{tabular}{c|cc|cc|cc}
\hline
\multirow{2}{*}{} & \multicolumn{2}{c|}{MSTAR} & \multicolumn{2}{c|}{SynthWakeSAR} & \multicolumn{2}{c}{GBSAR} \\ 
                  & VTR    & Multi-layer GNN  & VTR           & ResNet50         & VTR       & ResNet34      \\ \hline
\# Parameters     & \textbf{1.16M}   & 1.27M            & \textbf{1.37M}          & 23.5M            & \textbf{0.59M}      & 21.3M         \\ \hline
\end{tabular}
\end{table}

\section{CONCLUSION AND FUTURE WORK}
\label{sec:misc}
In this paper, we developed a lightweight \underline{V}iT model tailored for SAR A\underline{TR} applications, VTR, addressing the challenges of limited training data and model computational complexity. Our experimental results demonstrated the effectiveness of the proposed model across the diverse SAR datasets, achieving better or comparable results than prior work. The proposed FPGA accelerator is suitable for real-time SAR ATR workloads with tight latency constraints, performing significantly better than the alternative state-of-the-art GPU and CPU platforms. In future work, we will explore multi-modal datasets, such as the EO-SAR dataset, for the SAR ATR application and how such multi-modality can be exploited to improve the model performance. Furthermore, we will explore novel hybrid ViT and GNN architectures to overcome the performance limitations of purely ViT-based approaches on MSTAR-like data. To this extent, we will study approaches that exploit ViT's global and GNN's local inductive bias.
\\
 
\section{Acknowledgement}
This work is supported by the  DEVCOM  Army Research Lab (ARL) under grant W911NF2220159 and the National Science Foundation (NSF) under grants SPX-2333009 and SaTC-2104264. Equipment and support by AMD AECG are greatly appreciated. 
\newline
\textbf{Distribution Statement A}: Approved for public release. Distribution is unlimited.

\bibliography{report} 
\bibliographystyle{spiebib} 

\end{document}